\documentclass[sn-mathphys]{sn-jnl}

\usepackage{etoolbox}
    \makeatletter
    \patchcmd{\ps@headings}
    {\hbox to \hsize{\hfill Springer Nature 2021 \LaTeX\ template\hfill}}
    {\hbox to \hsize{\hfill Based on  Springer Nature \LaTeX\ template\hfill}}
    {}
    {}
    \patchcmd{\ps@titlepage}
    {\hbox to \hsize{\hfill Springer Nature 2021 \LaTeX\ template\hfill}}
    {\hbox to \hsize{\hfill Based on  Springer Nature \LaTeX\ template\hfill}}
    {}
    {}
    \makeatother

\jyear{2021}%

\theoremstyle{thmstyleone}%
\theoremstyle{thmstyletwo}%

\theoremstyle{thmstylethree}%

\usepackage{comment}
\usepackage{bm}

\begin{document}

\title[Revisiting $l_p$-constrained Softmax Loss]{Revisiting $l_p$-constrained Softmax Loss: A Comprehensive Study}


\author*{\fnm{Chintan} \sur{Trivedi$^*$}}\email{ctriv01@um.edu.mt}

\author{\fnm{Konstantinos} \sur{Makantasis}}\email{konstantinos.makantasis@um.edu.mt}

\author{\fnm{Antonios} \sur{Liapis}}\email{antonios.liapis@um.edu.mt}

\author{\fnm{Georgios} \sur{Yannakakis}}\email{georgios.yannakakis@um.edu.mt}

\affil{\orgdiv{Institute of Digital Games}, \orgname{University of Malta}, \orgaddress{\country{Malta}}}




\abstract{Normalization is a vital process for any machine learning task as it controls the properties of data and affects model performance at large. The impact of particular forms of normalization, however, has so far been investigated in limited domain-specific classification tasks and not in a general fashion. Motivated by the lack of such a comprehensive study, in this paper we investigate the performance of $l_p$-constrained softmax loss classifiers across different norm orders, magnitudes, and data dimensions in both proof-of-concept classification problems and real-world popular image classification tasks. Experimental results suggest collectively that $l_p$-constrained softmax loss classifiers not only can achieve more accurate classification results but, at the same time, appear to be less prone to overfitting. The core findings hold across the three popular deep learning architectures tested and eight datasets examined, and suggest that $l_p$ normalization is a recommended data representation practice for image classification in terms of performance and convergence, and against overfitting.}

\keywords{classification, normalization, deep learning, computer vision}



\maketitle

\section{Introduction}
\label{sec:intro}
Visual content representation in appropriate vector spaces is a fundamental task for any computer vision application \cite{marr1980visual}. 
Way before the era of deep learning, computer vision researchers predominately represented visual content by a set of handcrafted features exploiting the content's local and global properties as well as images' low-level information \cite{tuytelaars2008local,zheng2017sift,roth2008survey}. Although the handcrafted representations were well-defined and their statistical properties were known \emph{a priori}, they were computed independently of the subsequent processing and decision making steps. This gap of information flow---combined with no prior knowledge of which features are best suited for a specific computer vision task---often led to poor performance.

Thanks to the advancements of deep learning, and specifically Convolutional Neural Networks (ConvNets) and Vision Transformers (ViT), the aforementioned gap of information flow has been eliminated. ConvNets and ViT process visual information in its raw form (i.e. images) and learn to extract the most appropriate feature set for a given decision making task \cite{lecun1998gradient,sharif2014cnn}. The fact that deep learning automates the visual content representation task has led to breakthroughs in fundamental computer vision tasks such as image classification based on visual information \cite{krizhevsky2012imagenet}. The computed features, however, are depended on the parameterization of the employed deep learning architectures, the available training data, and the employed loss function. Therefore, the statistical properties, let alone the physical interpretation, of the features computed are barely known. Having no control over the properties of the features may lead to learning machines characterized by poor generalization capacity or misleading predictions \cite{yuan2019adversarial}. 

Several approaches have been proposed to control the properties of the computed features such as Batch Normalization \cite{ioffe2015batch}, Layer Normalization \cite{ba2016layer}, and $l_2$-constrained softmax loss \cite{tygert2015scale}. Although the first two are well-established and widely adopted techniques, the latter approach has received limited attention. To the best of our knowledge, $l_2$-constrained softmax loss has been introduced and applied only to a few domain-specific studies such as face verification \cite{ranjan2017l2,wang2017normface}, while its effect on the performance of learning machines has not been investigated comprehensively and in a general fashion. 

\begin{figure*}[t]
	\begin{minipage}{0.24\linewidth}
		\centering
		\centerline{\includegraphics[width=0.99\linewidth]{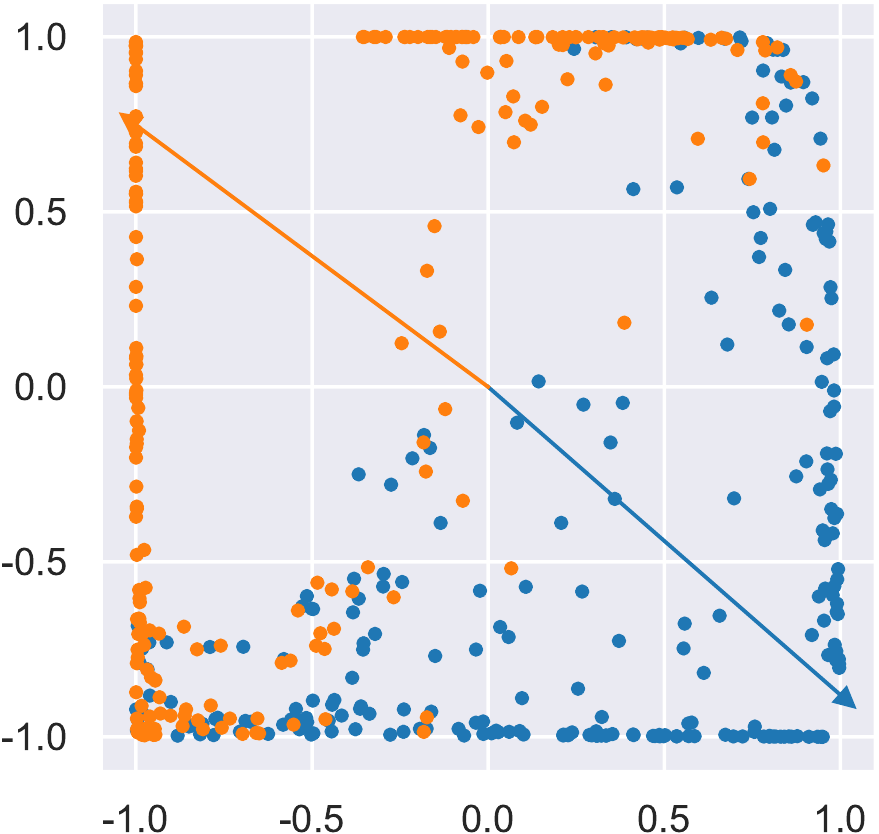}} \small{No normalization}
	\end{minipage}
	\begin{minipage}{0.24\linewidth}
		\centering
		\centerline{\includegraphics[width=0.99\linewidth]{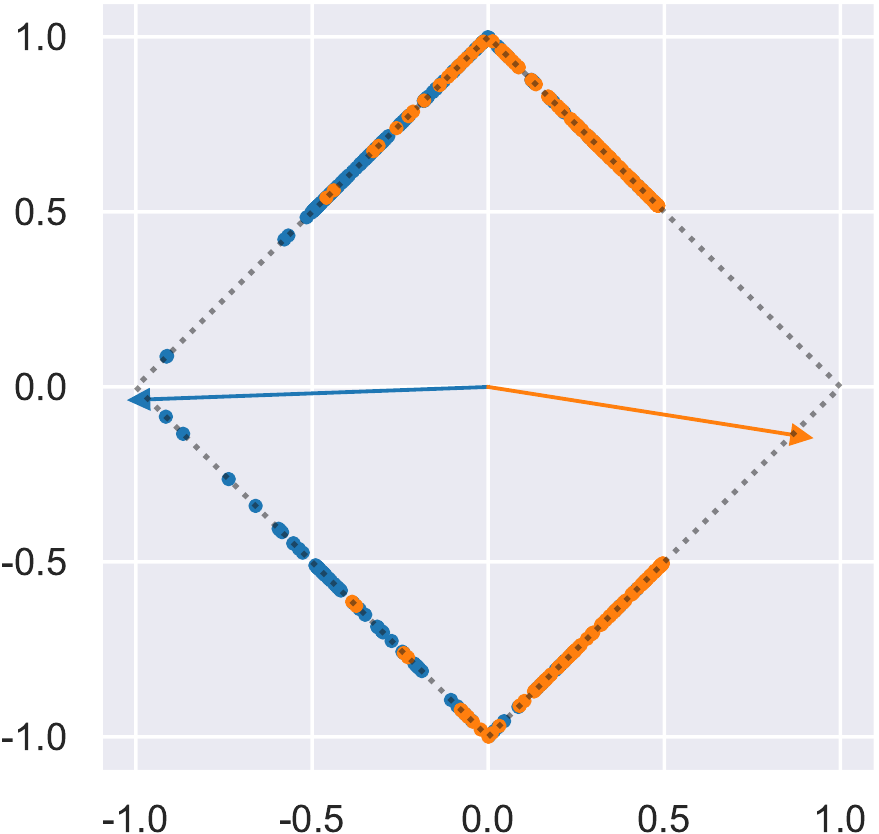}} \small{$\|\cdot\|_1$ normalization}
	\end{minipage}
	\begin{minipage}{0.24\linewidth}
		\centering
		\centerline{\includegraphics[width=0.99\linewidth]{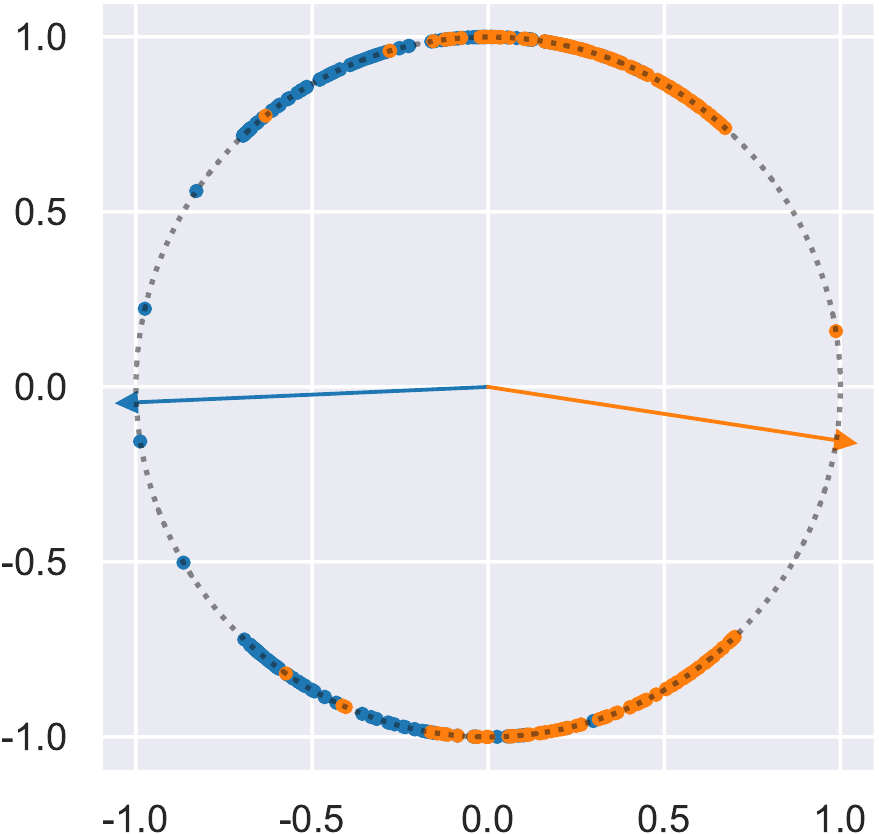}} \small{$\|\cdot\|_2$ normalization}
	\end{minipage}
	\begin{minipage}{0.24\linewidth}
		\centering
		\centerline{\includegraphics[width=0.99\linewidth]{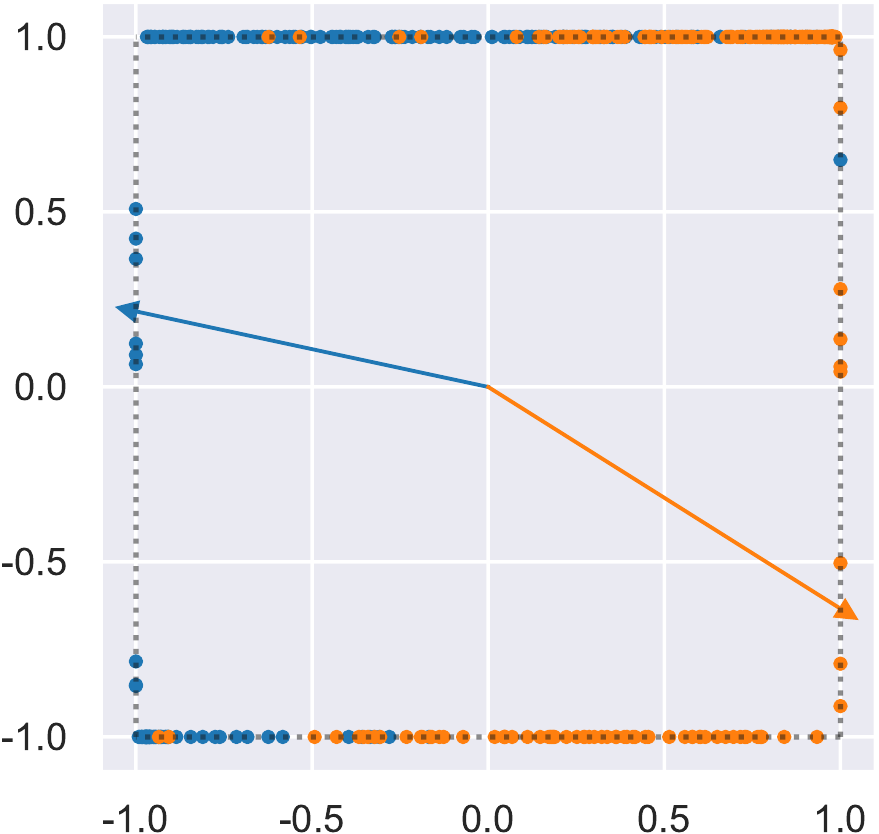}} \small{$\|\cdot\|_{\infty}$ normalization}
	\end{minipage}
	\vspace{0.05in}
	\caption{Projection of points belonging to two classes (orange or blue) on manifolds using norms of order 1, 2, and $\infty$ along with the weight vectors for each class (normalized for visualization purposes). Blue points occupy the first and third quadrants, while orange points the second and fourth. The points are classified using a neural network with two hidden layers with 16 and 2 neurons respectively and hyperbolic tangent activations. The points' representation is being projected on the manifolds is the 2D output of the second hidden layer.}
	\label{fig:projections} 
\end{figure*}

Motivated by the lack of a thorough study on $l_p$ normalization, this paper investigates the impact of $l_p$-constrained softmax loss---a generalization of $l_2$---on the performance of learning machines as applied to the general problem of image classification. The $l_p$-constrained softmax loss projects the feature representations of data points (e.g. images) on the surface of a manifold with fixed radius; i.e. the magnitude of the $l_p$ norm of the feature representations equals a fixed and predefined number. 

In this study, we investigate the effect of this projection on classification performance in terms of three core manifold properties: the order of the norm, the manifold dimension and its radius. Our evaluation builds on a proof-of-concept example that aims to shed light on the behaviour of $l_p$-constrained softmax loss, and proceeds with the application of $l_p$-constrained softmax loss on eight publicly available and widely-used image classification datasets. For the evaluation of $l_p$-constrained softmax loss we use three popular deep learning architectures: two ConvNet architectures and one ViT. Employment of $l_p$-constrained softmax loss consistently improves the performance of ConvNets. The ViT architecture, however, seems to benefit more when $l_p$-constrained softmax loss is used with datasets that consist of a large number of classes. Based on the outcomes of our experiments we can argue that $l_p$-constrained softmax loss can improve the performance of image classification, it leads to faster convergence and protects training against overfitting, and it can be used in conjunction with other normalization techniques; it is, thus, recommended as a general good machine learning practice in that domain.

\section{Background: Normalization} 

As mentioned earlier, the representations of data affect both the performance (generalization capacity) and the behaviour (convergence time, robustness to different initialization) of machine learning models. Therefore, controlling the properties of data representations is essential for successfully addressing pattern recognition tasks. In the following, we review core methods for controlling the properties of data representations that have been designed for and applied to neural network learning models. 

One of the first approaches for controlling the statistical properties of data representations has been proposed in \cite{ioffe2015batch} and is known as \emph{batch normalization}. As the name suggests, batch normalization standardizes the features (representation elements) across a batch of data points by removing the mean and scaling to unit variance. Batch normalization reduces convergence time, by reducing internal covariance shift \cite{santurkar2018does} and bounding the magnitude of the gradients, and improves networks performance \cite{ioffe2015batch}.   

In contrast to batch normalization, which standardizes the features across batches of data, the authors in \cite{ba2016layer} suggest standardizing data representations individually. This technique is known as \emph{layer normalization} and results in zero mean and unit variance representations. Layer normalization does not depend on the batch size, and in the case of recurrent neural networks, it yields more accurate networks than those featuring batch normalization \cite{xu2019understanding,vaswani2017attention}. 

In addition to batch and layer normalization, \emph{instance normalization} \cite{ulyanov2016instance} has explicitly been designed and proposed for computer vision tasks and ConvNets. It is applied on third-order tensor objects and standardizes them across two of the three dimensions, i.e. zero mean and unit variance tensor slices. This technique has been devised for style transfer applications \cite{huang2017arbitrary}, and usually it is applied along the spatial dimension of images to produce contrast-agnostic neural network models. A modification of instance normalization known as \emph{group normalization} \cite{wu2018group} is also used with third-order tensor objects: instead of standardizing tensor slices, as the name suggests, it standardizes groups of tensor slices. 

Instead of manipulating the statistical properties (mean and variance) of representations, another way to control their form is by requiring them to lie onto a specific manifold. Towards this direction, the studies in \cite{tygert2015scale, liu2017deep} force the data points' representations to lie on the surface of a unit hypersphere, i.e. the $l_2$ norm of representations equals one. This way, the representations' direction is retained, while their magnitude is always clipped to one yielding scale-invariant learning models. This approach, also known as \emph{$l_2$-constrained softmax loss}, has proved to be beneficial for a number of applications including face verification and identification \cite{liu2017sphereface,ranjan2017l2}, training auto-encoders \cite{davidson2018hyperspherical,DBLP:journals/corr/abs-1808-10805} and contrastive learning \cite{wang2020understanding}.

Batch, layer, instance and group normalization techniques are mainly used to stabilize training and improve convergence rate by reducing variance across batches of data points or individual instances as studied in \cite{huang2020normalization}. These techniques have been widely adopted by the research community and are ubiquitous in modern neural network applications. On the contrary, projecting the representations on the surface of a manifold has not been given the necessary attention yet as there are only limited studies proposing the use of this normalization type. In the current study, we aim to shed light on the effect of this technique on general image classification tasks in a comprehensive manner.

\begin{figure*}[tb]
	\begin{minipage}{1.0\linewidth}
		\centering
		\centerline{\includegraphics[width=0.97\linewidth]{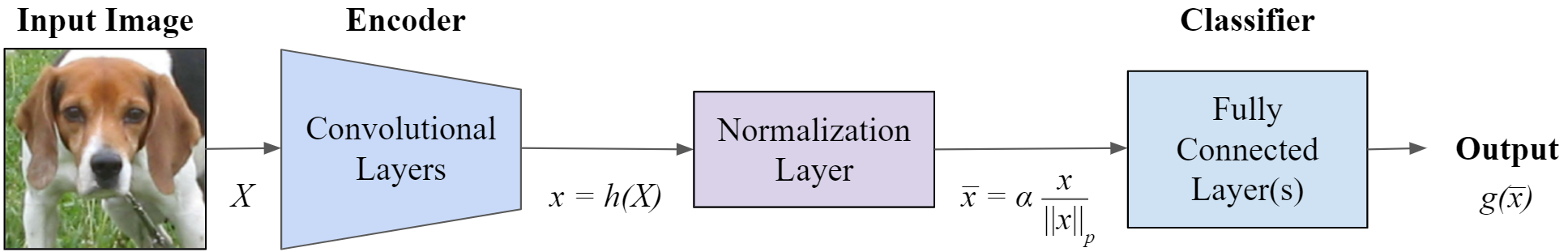}}
	\end{minipage}
	\vspace{0.05in}
	\caption{Information flow through a ConvNet that uses $l_p$-constrained softmax loss.}
	\label{fig:architecture} 
\end{figure*}

The novelty of this study is four-fold. First, we follow a thorough sensitivity analysis and investigate the behaviour of $l_p$-constrained softmax loss---a generalization of projecting representations in the surface of a hypersphere---in relation to the manifold's shape, dimension and radius. While $l_2$-constrained softmax loss has been studied in the literature sporadically, to the best of our knowledge this is the first time a study examines the effect of different norm orders (determining the shape of the manifold) on the performance of classifiers. Second, we quantify the impact of $l_p$-constrained softmax loss holistically: not only in terms of classification accuracy, but also in terms of classes' compactness, and in terms of its deviation from the optimal Bayes classifier. Third, we investigate the stability of $l_p$-constrained softmax loss classifiers on training and their robustness to overfitting via a proof-of-concept example that is easy to visualise and analyse. Finally, we conduct comprehensive experiments using eight widely-used datasets and three popular deep learning architectures to study the behaviour and impact of $l_p$-constrained softmax loss on real-world image classification tasks.

At this point we should mention that batch, layer and instance normalization can be applied to any layer of a neural network. On the contrary, $l_p$-constrained softmax loss operates only on the activations of the layer that produces the image representation to manipulate the manifold onto which the representation is projected. Therefore, we should stress out and clarify that $l_p$-constrained softmax loss can be used in conjunction and is complementary to other normalization techniques. In fact, in Section \ref{sec:results} we present the performance of $l_2$-constraned softmax loss classifiers when applied in conjunction with neural network architectures that employ batch and layer normalization.

The remainder of the paper is organized as follows: Section \ref{sec:lp_loss} briefly presents the mathematical formulation of $l_p$-constrained softmax loss classifiers, Section \ref{sec:experimental_framework} describes the experimental framework followed in this study. The experimental evaluation of $l_p$-constrained softmax loss classifiers is reported in Section \ref{sec:results}, and Section \ref{sec:conclusions} concludes this work.

\section{$l_p$-constrained Softmax Loss}\label{sec:lp_loss}

In typical image classification tasks, machine learning models are trained to classify a given image to its correct category. Let us denote as $\mathcal X \subset \mathbb R^{c \times h \times w}$ the input space of images with $c$ channels, height $h$ and width $w$, and as $\mathcal Y = \{1,2,\cdots,K\}$ the corresponding class categories. The objective of a machine learning model is to compute a function $f:\mathcal X \rightarrow \mathcal Y$ that maps an image $X \in \mathcal X$ to its corresponding category $y \in \mathcal Y$. 

In the case of ConvNets and ViT models the function $f$ can be seen as a composition of functions $h:\mathcal X \rightarrow \mathbb R^d$ and $g:\mathbb R^d \rightarrow \mathcal Y$. Function $h$ produces the feature representation of images, that is the output of model's penultimate layer, and function $g$ is the classifier that maps the representation $\bm x=h(X)$ of an image $X$ to a class category, i.e. 
\begin{equation}
	\label{eq:classifier}
	g(x) = \arg\max_{i \in \mathcal Y}(W_i^T h(X) + b_i),
\end{equation}
where $W_i$ is the $i$-th column of the model's output weights and $b_i$ the $i$-th element of the model's output bias vector. Both $W_i$ and $b_i$ are used to compute the output for the $i$-th class. By examining Eq.~\eqref{eq:classifier} it becomes clear that the predictions of a ConvNet or a ViT are determined by the value of 
\begin{equation}
	\label{eq:cos_distance}
	\langle W_i, h(X) \rangle = \|W_i\|_2 \cdot \|h(X)\|_2 \cdot \cos \theta, 
\end{equation}   
where $\theta$ is the angle between $W_i$ and $h(X)$ and $\|\cdot\|_2$ stands for the second norm of a vector. Therefore, the predictions are depended on the magnitudes of $W_i$ and $h(X)$. 

The parameters of the model presented above---that is the parameters of $f=g \circ h$---are estimated during training by minimizing the empirical softmax loss
\begin{equation}
	\label{eq:softmax}
	L =- \frac{1}{M}\sum_{m=1}^M \log \frac{\exp(W_{y_m}^T h(X_m) + b_i)}{\sum_{k=1}^K \exp(W_k^T h(X_m) + b_k)}
\end{equation}
over a training set $\mathcal D=\{(X_m, y_m)\}_{m=1}^M$ of $M$ image-category pairs.

The authors in \cite{ranjan2017l2} observed that training a model using the softmax loss function of Eq.~\eqref{eq:softmax} and letting $h(X)$ unrestricted encourages the network to learn high magnitude representations for well separated samples and ignore hard to classify samples, which deteriorates the generalization capacity of the classifier. This observation is also formalized in Proposition 1 in \cite{wang2017normface}. To avoid this drawback, the authors of \cite{ranjan2017l2,wang2017normface} normalize $h(X)$ such that its second norm is equal to a constant $\alpha$. 

In this work we study a more general normalization technique; specifically, we restrict the $p$-th norm of $h(X)$ to be equal to a constant $\alpha$. This way the normalized representation $\bar{\bm x}$ of an image $X$ is given by
\begin{equation}
	\label{eq:p_normalization}
	\bar{\bm x} = \alpha \frac{h(X)}{\|h(X)\|_p}.
\end{equation}
We investigate the impact of Eq.~\eqref{eq:p_normalization} on the performance of a classifier for different values of $p$ and $\alpha$, which can be fixed by the user or treated as learnable parameters that are being estimated during the training of the network. The value of $p$ in Eq.~\eqref{eq:p_normalization} affects, first, the shape of the manifold on which $h(X)$ is projected and, second, the value of $\|\bar{x}\|_2$ in Eq.~\eqref{eq:cos_distance}. Specifically, let 
\begin{equation}
    C_p=\frac{\|h(X)\|_2}{\|h(X)\|_p} ,
\end{equation}
then for different values of $p$, the value of $\|\bar{\bm x}\|_2$ lies in the following interval
\begin{equation}
	\label{eq:representation}
	\frac{\|\bar{\bm x}\|_2}{\alpha} \in \big[\min(1, C_p), \max(1, C_p)\big].
\end{equation}
Note that $C_2=1$ (when $p=2$) and thus $\|\bar{x}\|_2$ equals to $\alpha$.

Figure \ref{fig:projections} presents the projection of $h(\bm x)$ for $\bm x \in \mathbb R^2$ for $p=1, 2$ and $\infty$. The points depicted belong to one of two classes: blue and orange. Blue points occupy the first and third quadrants in a 2D Euclidean space and orange points the second and fourth. The points are classified using a neural network with two hidden layers with 16 and 2 neurons respectively, and hyperbolic tangent activations. The representation of points projected on manifolds is the 2D output of the networks' hidden layer. Figure \ref{fig:architecture}
presents the information flow for a ConvNet that employs this kind of normalization, which can be easily implemented using built-in functions from publicly available deep learning toolboxes such as Tensorflow \cite{abadi2016tensorflow}, Caffe \cite{jia2014caffe} and Pytorch \cite{paszke2019pytorch}. 

\section{Experimental Framework}\label{sec:experimental_framework}

In this section we present the experimental framework used for investigating the impact of $l_p$-constrained softmax loss on the performance of neural network classifiers. We start by introducing the employed datasets and models. Then, we define the experimental setting and the metrics used for evaluating models' performance. 

\subsection{Datasets}

We employ both synthetic and real-world datasets. Using synthetic data, we formulate a proof-of-concept problem for studying the robustness and generalization capacity of neural network classifiers that utilizes $l_p$-constrained softmax loss (see Section \ref{sec:pof}). Complementary to synthetic data we use eight different real-world datasets---corresponding to dissimilar image classification tasks--- and three popular deep learning architectures to evaluate the performance of $l_p$-constrained softmax loss classifiers in a comprehensive fashion (see Section \ref{sec:rwd}). 

\subsubsection{Proof-of-concept dataset} \label{sec:pof}

Instead of using only real-world datasets, we also investigate the behaviour of $l_p$-constrained softmax loss classifiers on a proof-of-concept problem using synthetic data. Through synthetic data we can both visualize the classifiers' decision boundary and test their generalization capacity by comparing them against the optimal Bayes classifier. The latter is possible since the Bayes classifier is defined in terms of the data generating distribution, which is known.

The synthetic dataset consists of $500$ two dimensional points that belong to two classes, i.e. binary classification in 2D Euclidean space. Each class consists of 250 points generated by a Mixture of two Gaussian components. For the first class the means of the components are $(-1.5, 1.5)$ and $(1.5, -1.5)$, while for the second class are $(-1.5, -1.5)$ and $(1.5, 1.5)$. All components have the same contribution---equal to 0.5---to the mixtures and their covariance is the diagonal matrix $\text{diag}([1.2, 1.2])$. According to this data generating distribution, all points of the first class are assigned to the first and third quadrants whereas and the points of the second class are placed onto the second and fourth. For this dataset, all models share the same architecture and their weights have been initialized using the same set of values. Moreover, we train all of them using the same set of data.

\begin{table}[!tb]
	\footnotesize
	\begin{center}
		\begin{tabular}{l@{ }||@{ }c@{ }|c|c|c|@{ }c@{ }}
			\hline\hline
			\multicolumn{1}{c||}{\textbf{Dataset}} & \textbf{Classes} & \textbf{Im. Size} & \textbf{Train} & \textbf{Val} & \multicolumn{1}{l}{\textbf{Batch}} \\ \hline\hline
			beans & 3 & 224x224 & 1034 & 133 & 64 \\ \hline
			cats vs dogs & 2 & 128x128 & 19,773 & 3,489 & 64 \\ \hline
			cifar10 & 10 & 32x32 & 50k & 10K & 128 \\ \hline
			cifar100 & 100 & 32x32 & 50k & 10K & 128 \\ \hline
			imagenette & 10 & 128x128 & 9,469 & 3,925 & 64 \\ \hline
			rock paper scissors & 3 & 224x224 & 2,520 & 372 & 32 \\ \hline
			stanford dogs & 120 & 128x128 & 12K & 8,580 & 64 \\ \hline
			tf flowers & 5 & 224x224 & 3,103 & 547 & 32 \\ \hline\hline
		\end{tabular}
	\end{center}
	\caption{Summary of datasets used and the respective training hyper-parameters.}
	\label{table:datasetsummary}
\end{table}

\subsubsection{Real-world datasets}\label{sec:rwd}

We evaluate the performance of $l_p$-constrained softmax classifiers across eight (8) publicly available datasets, see Table \ref{table:datasetsummary}, from the Tensorflow Datasets Catalog\footnote{www.tensorflow.org/datasets/catalog/overview\#image\_classification}. All eight datasets are widely used for benchmarking machine learning models on image classification tasks. 

The selection of the datasets to be examined is based on two core criteria. First, we choose to use datasets of varying \emph{size}; small-sized containing less than 10K images, medium-sized with 10K-25K images and large datasets with more than 25k images. Using varying size datasets, we can test the generalization capacity of the $l_p$-constrained softmax classifiers and their behaviour against overfitting. Second, we chose datasets with a varying number of \emph{classes}. As shown in \cite{bamler2019extreme}, softmax loss is affected by the number of classes due to the normalization factor (see the denominator in Eq.(\ref{eq:softmax})). Based on that, we selected datasets whose classes range from 2 to 120. This way, we can investigate the degree to which discriminating class weight vectors can be learnt on manifolds of specific shape and size when the number of classes changes. Moreover, we can measure the compactness of class-wise clusters formed on the representation space given that the normalized points (see Eq.(\ref{eq:representation})) should ideally form compact clusters based on their corresponding class labels. Finally, to increase the datasets' size, we used data augmentation techniques, including random horizontal flip, rotation and scaling of the images \cite{shorten2019survey}.

\subsection{Models}
\label{ssec:models}
For the proof-of-concept problem we use a fully connected neural network with two hidden layers and hyperbolic tangent activations. The $l_p$ normalization is applied to the output of the second hidden layer. The first hidden layer consists of 128 neurons, while the number of neurons at the second hidden layer ranges from 2 to 16 to test the effect of the representation's dimension on the model's performance.

For the real-world datasets, instead, we use the ResNet50 as an encoder that maps images to a $2,048$ dimension space. The ResNet50 encoder is followed by a dense layer that is responsible for carrying out the classification task. We use ResNet50 since it is a widely-used mid-sized ($\sim$25M parameters) ConvNet that achieves excellent performance in general image classification tasks. We train the encoder and the subsequent classification layer in an end-to-end fashion without using pre-trained weights to enable the network to learn appropriate $l_p$ normalized representations of the images. At this point, we should stress that we do not aim to produce state-of-the-art results on the datasets presented above. Instead, our aim is to examine whether the $l_p$ normalization can boost the performance of an image classifier.

\subsection{Experiments and Metrics}

For all datasets we run experiments investigating the effect of the $l_p$ normalization hyperparameters, that is $p$ and $\alpha$ in Eq.~\eqref{eq:representation}. These hyperparameters can be determined in advance, or become learnable parameters that are estimated during the classifier's training. Specifically, we test four different settings for parameter $p$ ($p=1,2,\infty$ and learnable) and two for parameter $\alpha$ ($\alpha=1$ and learnable).

For each hyperparameter setting in the real-world datasets, we run five separate experiments with different Glorot-Uniform weight initializations and random data augmentation to test the robustness of the models. We evaluate the effect of $l_p$ normalization on the performance of the models using a set of metrics. For the proof-of-concept problem---since the data generating distribution is known---we use the trained classifier's deviation from the Bayes optimal classifier; i.e. the probability that predictions of the trained classifier and the Bayes classifier are different when a point is randomly sampled from the data generating distribution. For the real-world datasets, we use the validation accuracy and the evolution of training loss. These are typical classification metrics, which also allow testing a classifier's behaviour in terms of overfitting. We also use the silhouette score \cite{rousseeuw1987silhouette} to quantify the compactness of class-wise clusters produced by $l_p$ normalized image representations. 

\section{Results}
\label{sec:results}

This section evaluates the impact of $l_p$-constrained softmax loss on models' performance following the experimental framework described earlier in Section \ref{sec:experimental_framework}. We start by presenting the results for the proof-of-concept problem and move on with the results obtained on the real-world data\footnote{The code is available at https://github.com/ChintanTrivedi/LpSoftmaxLoss}.

\subsection{Proof-of-concept problem}

\begin{figure}[!tb]
	\begin{minipage}{0.99\linewidth}
		\centering
		\centerline{\includegraphics[width=0.99\linewidth]{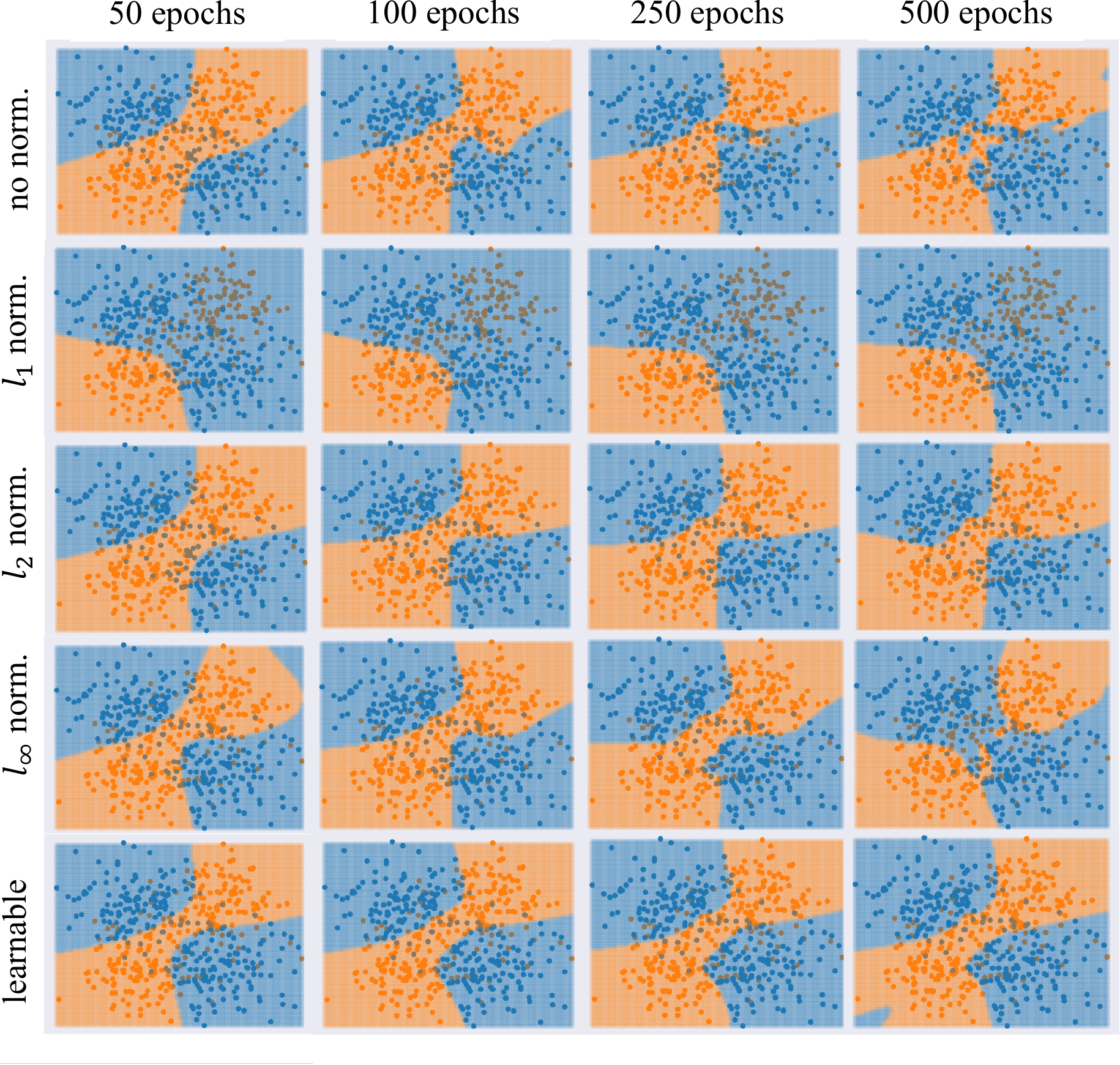}}
	\end{minipage}
	\caption{Decision boundaries for classifiers using $l_p$ normalization on the proof-of-concept data.}
	\label{fig:decision_boundary}
\end{figure}

Running experiments using the proof-of-concept data and models described in Section \ref{sec:experimental_framework} allows us to investigate the behaviour of $l_p$-constrained softmax loss classifiers in terms of their decision boundaries as well as their deviation from the optimal Bayes classifier.  

Figure \ref{fig:decision_boundary} presents the decision boundaries for different parameterized $l_p$-constrained softmax loss classifiers and the boundary for a classifier that does not use any normalization. For these results, we use the model described in Section \ref{sec:experimental_framework} with two neurons at its penultimate layer---we apply $l_p$ normalization to 2D data representations.

The classifier that does not use any normalization produces good decision boundaries when the training time is appropriately set; i.e. $100$ epochs. If we over-train the classifier, however, it starts overfitting the data after 250 epochs and it becomes more apparent at 500 training epochs. On the contrary, the decision boundary of $l_p$ constrained softmax loss classifiers, when $p$ is learnable and $p=2$, does not seem to be affected by over-training. These models learn accurate decision boundaries and retain them independently of training time. The classifier that uses $l_{\infty}$ normalization behaves similar to the classifier that does not use any normalization; its decision boundary is accurate, but it is not robust to over-training. Finally, the classifier that employs $l_1$ normalization underfits the data regardless of the number of training epochs used. These results suggest that $l_p$ constrained softmax loss classifiers can learn accurate decision boundaries, and at the same time, they are robust against overfitting. However, they have to be appropriately parameterized to avoid local minima and underfitting. 

\begin{figure}[!tb]
	\begin{minipage}{1.0\linewidth}
		\centering
		\centerline{\includegraphics[trim={0.0cm 0.0cm 0.0cm 0.0cm},clip,width=1.0\linewidth]{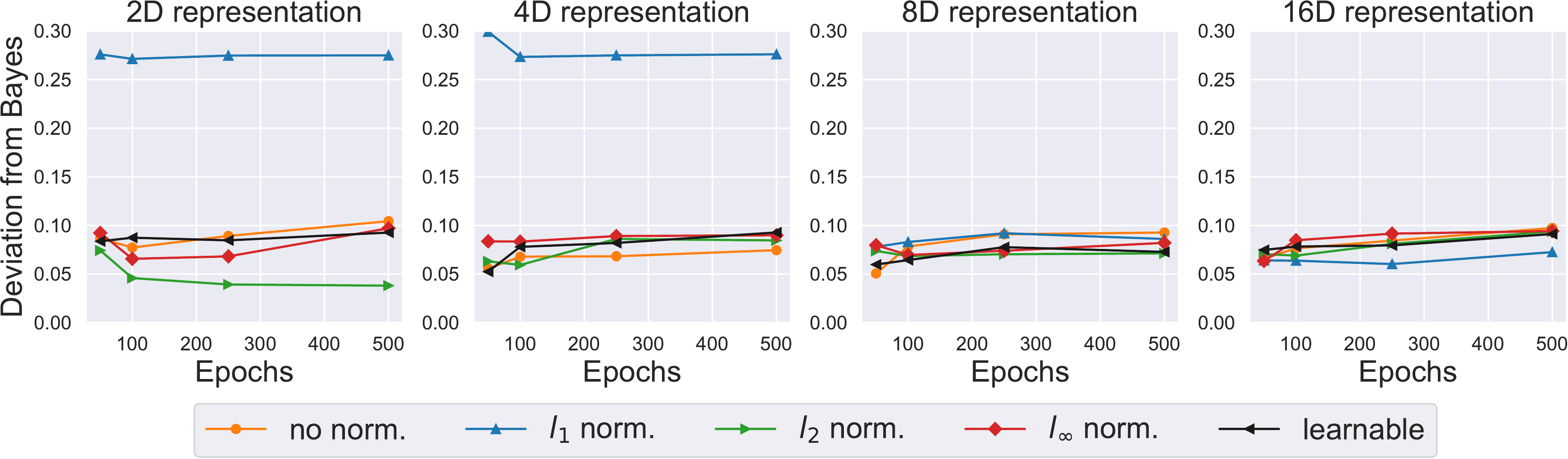}}
	\end{minipage}
	\vspace{0.05in}
	\caption{Deviation from the optimal Bayes classifier.}
	\label{fig:bayes}
\end{figure}

In Figure \ref{fig:bayes} we evaluate more formally the classifiers' performance by measuring their deviation from the optimal Bayes classifier. We also try to relate the behaviour of  $l_p$ normalization to the dimension of data representations. Towards this direction, we apply $l_p$ normalization to data representations of varying dimension, i.e. the number of neurons at the penultimate layer of the classifiers is 2, 4, 8, and 16. Deviation from the optimal Bayes decision rule quantifies the models' generalization capacity; low deviation indicates that a model can produce accurate predictions on unseen data, while high deviation suggests that a model either overfits or underfits the data.

The left diagram in Figure \ref{fig:bayes} corresponds to the decision boundaries presented in Figure \ref{fig:decision_boundary}. The $l_2$-constrained softmax loss classifier can learn the most accurate decision boundary without being affected by over-training. All $l_p$-constrained softmax loss classifiers, except the one that uses $l_1$ normalization, produce better decision rules than the classifier that does not use any normalization. The same holds for 8-dimensional and 16-dimensional data representations, but, surprisingly, not for the 4-dimensional ones. In this case, the unconstrained classifier that does not use any normalization technique outperforms all other classifiers after 150 epochs of training. Despite that result, we can conclude that the performance of $l_p$-constrained softmax loss classifiers affected less by the dimension of data representation.

A surprising result regards the behaviour of the $l_1$-constrained softmax loss classifier. For small data representations, it highly underfits the data. When the dimension increases, however, it produces comparable or even better results compared to the other classifiers. The fact that $l_1$ normalization produces better results with overparameterized models is interesting and needs to be further investigated both experimentally and theoretically. 

To summarize, results from the proof-of-concept problem suggest that $l_p$-constrained softmax loss classifiers---when they are appropriately set in terms of parameter $p$---are more robust to overfitting and improve classification performance. 

\subsection{Real-world Datasets}

In this section we investigate the behaviour of $l_p$-constrained softmax loss, in terms of its parameters, i.e. $p$, $\alpha$ and representation dimension, using real-world data.

\subsubsection{Varying Orders of Norm ($p$)}

Parameter $p$ corresponds to the order of the norm that is used for normalizing data representations. Following the procedure presented in Section \ref{sec:experimental_framework}, we evaluate classifiers' performance when $p$ is fixed to $1, 2$ and $\infty$, and when $p$ is a learnable parameter. We compare their performance against a classifier that is not using any normalization technique. Throughout this set of experiments, we use the ResNet model presented in Section \ref{ssec:models}. 

\begin{figure}[!tb]
	\centering
	\includegraphics[width=\linewidth]{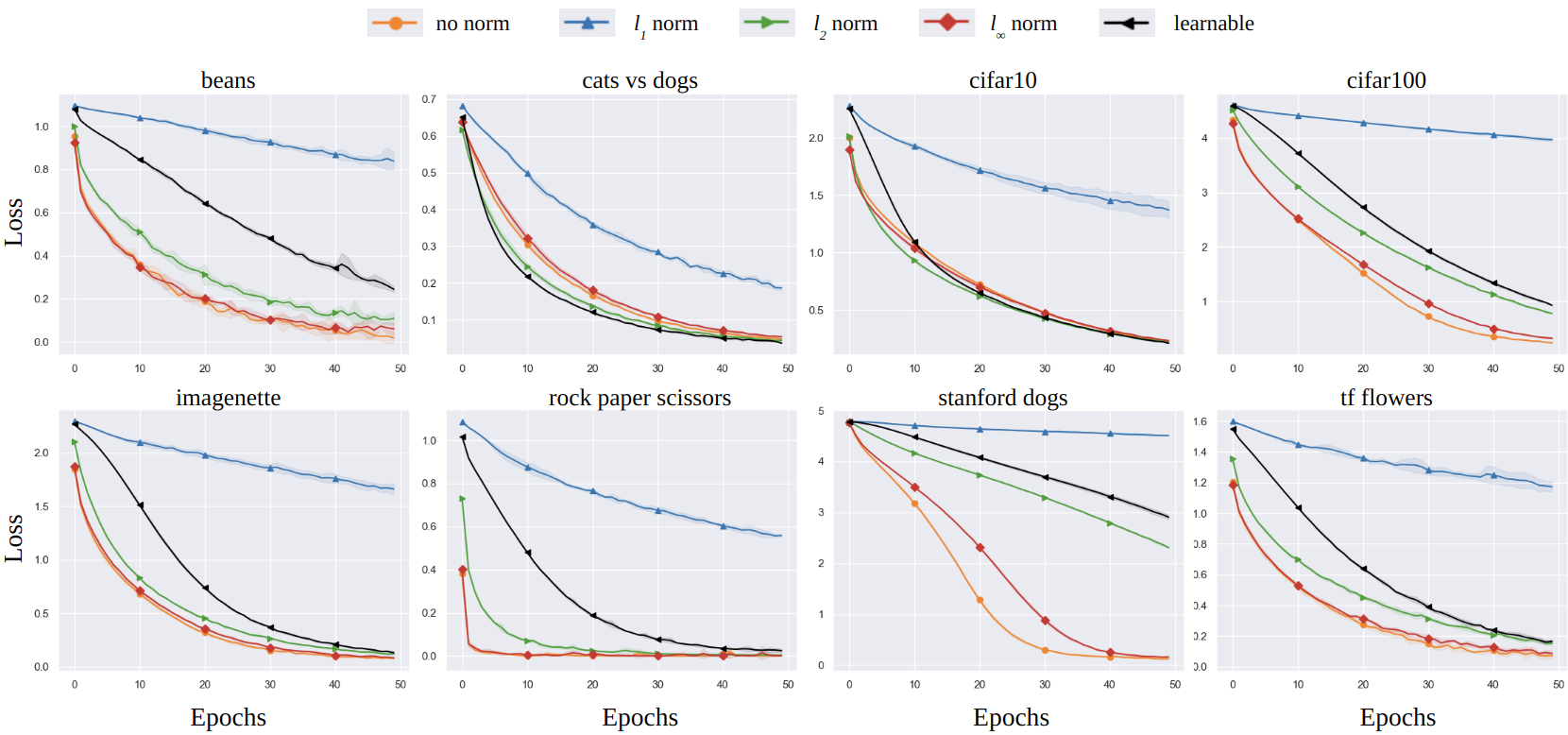}
	\caption{Progress of training loss for different values of $p$.}
	\label{fig:parameter_p}
\end{figure}

\begin{table*}[t]
	\footnotesize
	\begin{center}
		\begin{tabular}{ l||c|c|c|c|c} 
			\hline\hline
			\multicolumn{1}{c||}{\textbf{Dataset}} & \textbf{No norm} & \textbf{$l_1$ norm} & \textbf{$l_2$ norm} & \textbf{$l_{\infty}$ norm} & \textbf{learnable} \\ \hline \hline 
			beans &  81.7 $\pm$ 2.3 & 65.6 $\pm$ 17.5 & \textbf{84.7 $\pm$ 6.8} &  75.8 $\pm$ 4.7 & 82.0 $\pm$ 9.5 \\ \hline
			cats-vs-dogs & 90.2 $\pm$ 0.9 & 91.1 $\pm$ 1.5 & 90.9 $\pm$ 1.6 & 89.9 $\pm$ 3.2 &	\textbf{91.1 $\pm$ 0.9} \\ \hline
			cifar10 & 69.6 $\pm$ 0.5 & 41.0 $\pm$ 4.8 & 71.3 $\pm$ 0.5 & 69.9 $\pm$ 0.7 & \textbf{72.6 $\pm$ 1.3} \\ \hline
			cifar100 & 35.6 $\pm$ 1.0 & 5.5 $\pm$ 0.9 & \textbf{37.6 $\pm$ 0.7} & 36.9 $\pm$ 0.3 & 36.2 $\pm$ 0.7 \\ \hline
			imagenette & 64.6 $\pm$ 1.7 & 43.9 $\pm$ 4.9 & 67.8 $\pm$ 2.1 & 61.2 $\pm$ 4.5 & \textbf{69.0 $\pm$ 2.2} \\ \hline
			rock-paper-scissors & 86.3 $\pm$ 6.3 & 81.1 $\pm$ 22.9 & 96.5 $\pm$ 5.7 & 87.0 $\pm$ 18.6 & \textbf{99.3 $\pm$ 1.3} \\ \hline
			stanford-dogs & 11.3 $\pm$ 1.4 & 2.8 $\pm$ 0.7 & 11.4 $\pm$ 1.2 & \textbf{11.8 $\pm$ 1.1} & 8.7 $\pm$ 0.6 \\ \hline
			tf-flowers & 73.8 $\pm$ 3.2 & 63.4 $\pm$ 3.0 & 76.5 $\pm$ 2.8 & 70.7 $\pm$ 5.3 & \textbf{78.6 $\pm$ 3.4} \\ 
			\hline  \hline
		\end{tabular}
	\end{center} 
	\caption{Validation set classification accuracy and corresponding $95\%$ confidence intervals across 5 independent trials for different values of parameter $p$. Values in bold indicate the highest average accuracy obtained within a dataset.}
	\label{table:parameter_p}
\end{table*}

Figure \ref{fig:parameter_p} presents how training loss changes in the first 50 epochs of training. Solid lines represent the mean loss for five runs and shaded areas show the $95\%$ confidence interval. For all datasets, except \emph{stanford-dogs}, the classifiers converge within 50 training epochs. All classifiers, except $l_1$-constrained softmax loss, converge within 50 epochs of training for six out of eight datasets. For the \emph{stanford-dogs} and \emph{cifar100} datasets, the classifiers need more epochs to converge as these datasets contain many classes and can be considered harder as a machine learning task. As shown in Figure \ref{fig:parameter_p}, $l_p$-constrained softmax loss classifiers that have their $p$ parameter appropriately set ($\alpha=1$ for all experiments) converge faster and, in many cases, to lower loss values than the classifier that is not using any normalization technique. The worst performing $l_p$-constrained softmax loss classifier is the one that employs $l_1$ normalization. Actually, $l_1$-constrained softmax loss classifier does not manage to converge within 50 epochs of training for none of the eight employed datasets. On the contrary, the classifier that sets $p=\infty$ converges faster than all the other $l_p$-constrained softmax loss classifiers.

Besides convergence rates, we also investigate the classification accuracy on the validation set (unseen during training), since this is a far more representative metric for evaluating a classifier's performance. Table \ref{table:parameter_p} presents the mean validation accuracies and the corresponding $95\%$ confidence intervals of the models. $l_p$-constrained softmax loss classifiers generally perform better in most cases than the baseline classifier with no normalization. Specifically, $l_2$-constrained softmax loss classifier outperforms the baseline classifier that uses no normalization in all datasets, while in 2 (\emph{cifar10} and \emph{cifar100}) out of 8 datasets its accuracy is significantly higher (based on 95\% confidence intervals) than the baseline. The $l_\infty$-constrained softmax loss classifier outperforms the baseline classifier in 4 out of 8 datasets, and in \emph{cifar100} it performs significantly better than the baseline. When we treat the $p$ parameter as a learnable one, the $l_p$-constrained softmax loss classifier performs better than the baseline in 7 out of 8 datasets and in 3 of them (\emph{cifar10}, \emph{imagenette} and \emph{roch-paper-scissors}) significantly better. Finally, $l_1$-constrained softmax loss classifier acheives better performance than the baseline classifier in only one dataset (\emph{cats-vs-dogs}).

Based on the results obtained so far, we can safely argue that $l_p$ normalization for $p=2$ or learnable $p$ can boost an image classifier's performance. More importantly, this performance improvement comes for free, without any sophisticated or ambiguous model design choices.


\subsubsection{Varying Manifold Scale ($\alpha$)}

In this section we investigate the degree to which the radius (parameter $\alpha$) of the manifold on which the point representations are projected affects the performance of an $l_p$-constrained softmax loss classifier. In this set of experiments we again use the ResNet model described in Section \ref{ssec:models}. We run experiments for fixed $\alpha=1$ and for learnable $\alpha$, and we use two different classifiers: one that employs $l_2$ normalization and one the treats $p$ as a learnable parameter. The performance evaluation takes place in terms of classification accuracy on validation sets. 

\begin{figure}[!tb]
	\centering
	\includegraphics[width=\linewidth]{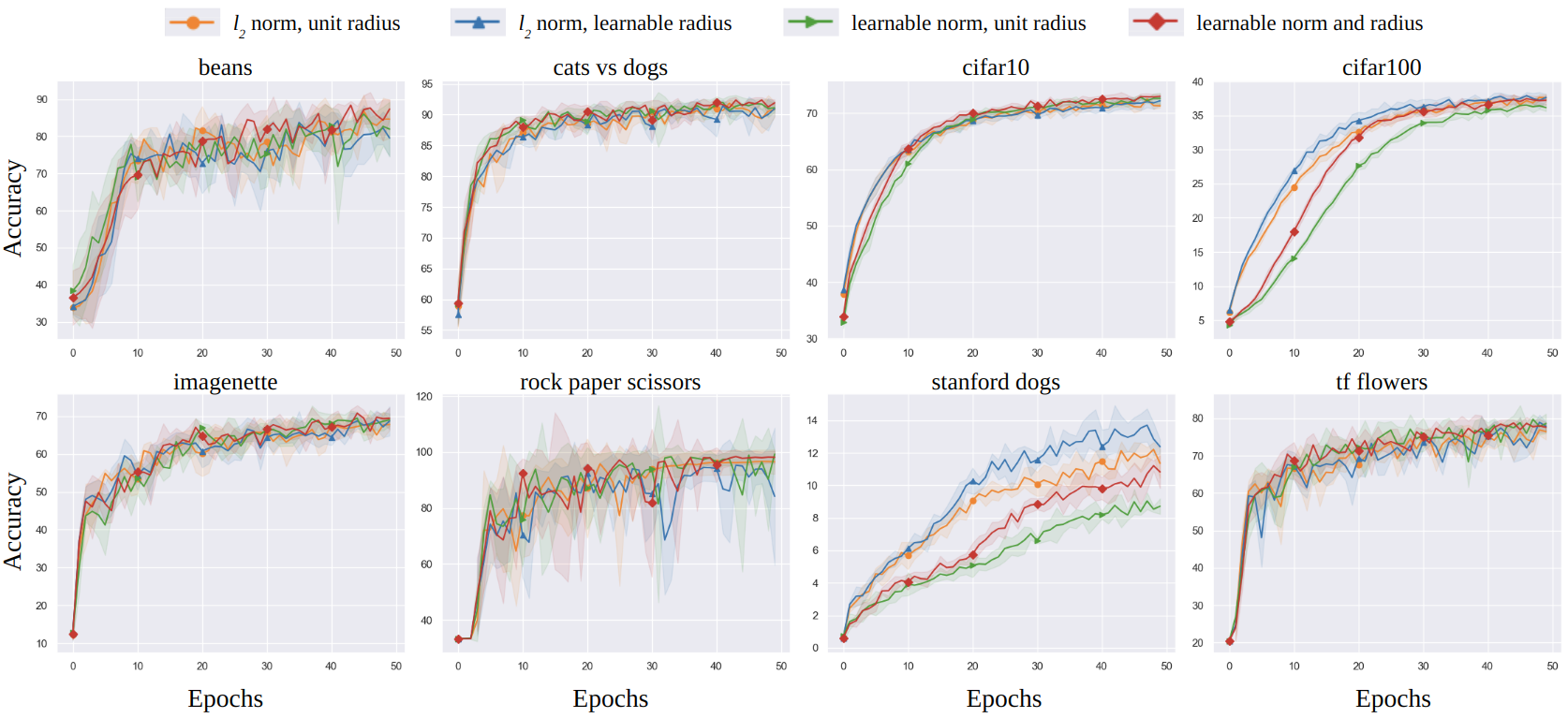}
	\caption{Progress of validation accuracy for different values of $\alpha$.}
	\label{fig:parameter_a}
\end{figure}

Figure \ref{fig:parameter_a} presents the results for this investigation for the two classifiers. While there are some fluctuations in early stages of learning, for 7 out of the 8 employed datasets the final accuracy of the models seems unaffected by the tested $\alpha, p$ parameter sets. For the \emph{stanford-dogs} dataset, however, the classifier that uses $l_2$ normalization and learnable $\alpha$ seems to achieve significantly higher accuracy compared to the other parameters settings. 

The above finding is in alignment with the results obtained in \cite{ranjan2017l2,wang2017normface} where the authors relate the parameter $\alpha$ with the number of available classes: the larger the number of classes, the larger the value of $\alpha$ should be. Based on these results, we can conclude that allowing $\alpha$ to be estimated during the model's training is a safe choice no matter what the number of classes is.

\subsubsection{Varying Dimensions of Representation}

In the experiments reported thus far, the ResNet model described in Section \ref{ssec:models} maps images to 2048 dimension vector representations, which then are normalized and propagated to the network's output via a hidden layer i.e. the classification boundary is non-linear in the space of the normalized representations. 

In the following, we investigate how the linearity/non-linearity and complexity--in terms of the number of trainable parameters--of classifiers affect the layout of the learned representations on the manifold. In other words, we investigate the compactness and separability of the classes to which the normalized data representations belong. 

We use dense neural networks with one hidden layer and vary the number of hidden neurons to investigate $l_p$ normalization when applied to representations of different dimensions. The quality of normalized representations is evaluated in terms of silhouette score as this metric quantifies the compactness of the classes and their separability on the manifold.

Table \ref{table:silhouette} presents the silhouette score for nonlinear and linear classifiers. For $l_p$ normalization we used $p=2$ and $\alpha=1$ to assure that the radius and the form of the manifold on which the representations are projected are the same for all classifiers. This way, the computed silhouette scores are directly comparable. For all datasets examined, except the \emph{stanford-dogs}, the linear classifier produces the highest silhouette score, on average. Specifically, its silhouette score is significantly higher than when using a nonlinear classifier with 4096 features in 4 datasets, with 2048 features in 5 datasets, with 512 features in 3 datasets, and with 128 features in 1 dataset.

As far as the nonlinear classifier is concerned, we observe that the silhouette score monotonically decreases as we increase the representation's dimension. For \emph{cifar100} and \emph{stanford-dogs} datasets, however, all classifiers perform equally poorly and the classes overlap in the representation space due to large number of classes. These findings suggest that the simpler the classifier, the better (or at least equal) the quality of the normalized representations on average, in terms of class compactness and separability. 

\begin{table*}[t]
	\footnotesize
	\begin{center}
		\begin{tabular}{l||c|c|c|c|c}
			\hline\hline
			\multicolumn{1}{c||}{\multirow{2}{*}{\textbf{Dataset}}} & \multicolumn{5}{c}{\textbf{Feature Representation Size}} \\ 
			\multicolumn{1}{c||}{} & \textbf{4096} & \textbf{2048} & \textbf{512} & \textbf{128} & \textbf{0} \\ \hline \hline
			beans & 	0.36$\pm$0.06 & 	0.38$\pm$0.02 & 	0.41$\pm$0.06 & 	0.44$\pm$0.04 & 	\textbf{0.49$\pm$0.10} 	\\ \hline
			cats-vs-dogs & 	0.46$\pm$0.03 & 	0.46$\pm$0.01 & 	0.48$\pm$0.01 & 	0.48$\pm$0.02 & 	\textbf{0.54$\pm$0.02} \\ \hline
			cifar10 & 	0.11$\pm$0.00 & 	0.12$\pm$0.00 & 	0.14$\pm$0.00 & 	0.16$\pm$0.01 & 	\textbf{0.17$\pm$0.00} 	\\ \hline
			cifar100 & -0.12$\pm$0.00 & -0.12$\pm$0.00 & -0.13$\pm$0.00 & -0.14$\pm$0.00 & \textbf{-0.11$\pm$0.00} \\ \hline
			imagenette & 	0.11$\pm$0.02 & 	0.13$\pm$0.00 & 	0.14$\pm$0.01 & 	0.16$\pm$0.03 & 	\textbf{0.20$\pm$0.02} 	\\ \hline
			rock-paper-scissors & 	0.69$\pm$0.09 & 	0.63$\pm$0.11 & 	0.78$\pm$0.05 & 	0.80$\pm$0.07 & 	\textbf{0.86$\pm$0.09} 	\\ \hline
			stanford-dogs & 	-0.11$\pm$0.01 & 	\textbf{-0.11$\pm$0.00} & 	-0.12$\pm$0.00 & 	-0.14$\pm$0.01 & 	-0.14$\pm$0.01 \\ \hline
			tf-flowers & 	0.20$\pm$0.02 & 	0.23$\pm$0.01 & 	0.23$\pm$0.03 & 	0.26$\pm$0.03 & 	\textbf{0.30$\pm$0.04} 	\\
			\hline \hline
		\end{tabular}
	\end{center}
	\caption{Average silhouette scores after 50 epochs for different representation sizes of the penultimate layer. Values are averaged across 5 runs with $95\%$ confidence intervals. Values in bold indicate the highest average silhouette score obtained within a dataset.
	}
	\label{table:silhouette}
\end{table*}

\subsubsection{$l_p$ norm normalization and other normalization techniques}

In the previous sections, we investigated the behaviour of $l_p$-constrained softmax loss classifiers in terms of their parameters, that is parameter $p$ representing the norm order and parameter $\alpha$ representing the radius of the manifold on which the data representations are projected. In this section we investigate the degree to which $l_p$-constrained softmax loss classifiers can be used in conjunction with other data normalization techniques. Specifically, we investigate the behaviour of $l_p$ norm normalization when the classifier employs batch normalization and layer normalization. 

To investigate the behaviour of $l_p$ norm normalization in conjunction with batch normalization we use the EfficientNet-B0 \cite{tan2019efficientnet} architecture, since this model employs batch normalization. While batch normalization is widely used with ConvNets, layer normalization is used with models that take into consideration sequential data. For this reason, we investigate the behaviour of $l_p$ norm normalization using a Vision Transformer (ViT) model, since this model is used for image classification tasks exploiting sequential data and employs layer normalization. Throughout these experiments we set parameter $p$ of the $l_p$-constrained softmax loss classifiers to 2.

Table \ref{table:Eff_ViT} presents the classification accuracy on validation sets and corresponding $95\%$ confidence intervals for EfficientNet-B0 and ViT. We comapre the performance of the models with and without applying $l_2$ norm normalization. In the case on EfficientNet-B0, which is a popular ConvNet architecture the classification performance improves for 7 out of the 8 datasets and significantly better (based on 95\% confidence interval) for 2 of them (\emph{cifar10}, \emph{stanford-dogs}). Therefore, based on these results we conclude that $l_2$ normalization can be used in conjunction with batch normalization to improve the performance of a ConvNet.

In the case of ViT the results are more balanced. The ViT that employs $l_2$ norm normalization performs better for 4 out of the 8 datasets, while for 2 of them (\emph{cifar100} and \emph{stanford-dogs}) it performs significantly better. However, similarly, the ViT that does not apply $l_2$ norm normalization on data representations performs better for 4 out of the 8 datasets, and for 2 of them (\emph{rock-paper-scissors} and \emph{cifar10}) significantly better. Based on these results, we can conclude that in the case of ViT, which employs layer normalization, $l_2$ norm normalization is beneficial when the number of classes is large. Both \emph{cifar100} and \emph{stanford-dogs} datasests for which the ViT with $l_2$ norm normalization performs significantly better include data from 100 and 120 classes, respectively.    

\begin{table*}[t]
	\footnotesize
	\begin{center}
        \begin{tabular}{l||cc|cc}
        \hline \hline
        \multicolumn{1}{c||}{\multirow{2}{*}{\textbf{Dataset}}} & \multicolumn{2}{c|}{\textbf{EfficientNet-B0}} & \multicolumn{2}{c}{\textbf{Visual Transformer}} \\ 
        \multicolumn{1}{c||}{} & \multicolumn{1}{c|}{\textbf{No Norm}} & \textbf{$l_2$ Norm} & \multicolumn{1}{c|}{\textbf{No Norm}} & \textbf{$l_2$ Norm} \\ \hline \hline
        beans & \multicolumn{1}{c|}{ \textbf{76.4 $\pm$ 3.9}} & 75.5 $\pm$ 5.2 & \multicolumn{1}{c|}{80.5 $\pm$ 1.1} & \textbf{82.7 $\pm$ 2.4} \\ \hline
        cats-vs-dogs & \multicolumn{1}{c|}{85.5 $\pm$ 1.2} & \textbf{86.7 $\pm$ 0.9} & \multicolumn{1}{c|}{\textbf{79.5 $\pm$ 0.7}} & 79.0 $\pm$ 1.9 \\ \hline
        cifar10 & \multicolumn{1}{c|}{82.0 $\pm$ 0.2} & \textbf{83.1 $\pm$ 0.7} & \multicolumn{1}{c|}{\textbf{76.1 $\pm$ 0.7}} & 73.6 $\pm$ 0.8 \\ \hline
        cifar100 & \multicolumn{1}{c|}{50.2 $\pm$ 0.5} & \textbf{51.3 $\pm$ 0.7} & \multicolumn{1}{c|}{46.6 $\pm$ 0.9} & \textbf{50.7 $\pm$ 0.9} \\ \hline
        imagenette & \multicolumn{1}{c|}{62.3 $\pm$ 3.3} & \textbf{64.0 $\pm$ 1.6} & \multicolumn{1}{c|}{\textbf{67.7 $\pm$ 0.5}} & 65.9 $\pm$ 2.1 \\ \hline
        rock-paper-scissors & \multicolumn{1}{c|}{80.1 $\pm$ 5.1} & \textbf{83.3 $\pm$ 6.9} & \multicolumn{1}{c|}{\textbf{83.7 $\pm$ 5.0}} & 65.7 $\pm$ 6.0 \\ \hline
        stanford-dogs & \multicolumn{1}{c|}{9.1 $\pm$ 0.3} & \textbf{11.0 $\pm$ 0.4} & \multicolumn{1}{c|}{10.3 $\pm$ 0.5} & \textbf{12.0 $\pm$ 0.1} \\ \hline
        tf-flowers & \multicolumn{1}{c|}{74.3 $\pm$ 2.2} & \textbf{74.7 $\pm$ 2.7} & \multicolumn{1}{c|}{69.4 $\pm$ 1.7} & \textbf{71.5 $\pm$ 1.1} \\ \hline \hline
        \end{tabular}
	\end{center}
	\caption{Validation set classification accuracy and corresponding $95\%$ confidence intervals across 5 independent trials for EfficientNet-B0 and ViT. Values in bold indicate the highest average accuracy obtained within a dataset.}
	\label{table:Eff_ViT}
\end{table*}

\section{Conclusions} 
\label{sec:conclusions}

This study evaluates the behaviour of $l_p$-constrained softmax loss classifiers on a proof-of-concept problem and eight popular real-world image classification tasks. Our experimental results across all tasks examined suggest that $l_p$ normalization---when appropriately parameterized---improves classification performance, leads to faster convergence and shields models against overfitting. Using the second norm ($p=2$) or treating $p$ as a learnable parameter for normalizing the data appear to produce the most robust results across all experimental settings. The size of the manifold---i.e. parameter $\alpha$ controlling the magnitude of normalized data representations---can affect the classifier's performance when the employed dataset consists of many classes; this hyperparameter, however, can be estimated during training time. Of particular interest is the behaviour of $l_1$ normalization when employed on overparameterized data representations. While overparameterization boosts the capacity of learning models, it also increases the risk for overfitting; $l_1$ normalization, however, seems to permit the increase in learning capacity without letting the models to overfit. This behaviour needs to be further investigated both experimentally and theoretically. Finally, $l_p$-constrained softmax loss consistently improves the performance of ConvNet architectures, while the benefits for ViT models are more obvious when the classification task employs a large number of classes.


To conclude, the key findings of this study suggest that $l_p$ normalization improves classification performance in a general fashion, in conjunction with other normalization techniques and for different model architectures by producing classifiers less prone to overfitting. Importantly, $l_p$ normalization can be used without any sophisticated or ambiguous classifier design choices and can be easily implemented using built-in functions from publicly available deep learning toolboxes. The limited set of hyperparameters that define $l_p$ normalization can also be learnable without a loss in performance.

\bmhead{Acknowledgments}
This work has been supported by the European Union’s Horizon 2020 research and innovation programme from the TAMED project (Grant Agreement No. 101003397).

\bibliography{egbib}


\end{document}